\documentclass[runningheads]{llncs}
\usepackage{graphicx}
\usepackage{comment}
\usepackage{amsmath,amssymb} 
\usepackage{color}

\begin{document}
\pagestyle{headings}
\mainmatter
\def\ECCVSubNumber{975}  

\title{Certainty Pooling for Multiple Instance Learning} 

\titlerunning{Certainty Pooling for Multiple Instance Learning}
%
\author{Jacob Gildenblat\inst{1} \and
Ido Ben-Shaul\inst{1} \and
Zvi Lapp\inst{1} \and
Eldad Klaiman\inst{2}}
\authorrunning{J. Gildenblat et al.}
%
\institute{SagivTech Ltd., Israel
\email{\{jacob,ido,zvi\}@sagivtech.com}\\
\and
Roche Innovation Center Munich, Germany\\
\email{eldad.klaiman@roche.com}}
\maketitle

\begin{abstract}
Multiple Instance Learning is a form of weakly supervised learning in which the data is arranged in sets of instances called bags with one label assigned per bag. The bag level class prediction is derived from the multiple instances through application of a permutation invariant pooling operator on instance predictions or embeddings. We present a novel pooling operator called \textbf{Certainty Pooling} which incorporates the model certainty into bag predictions resulting in a more robust and explainable model. We compare our proposed method with other pooling operators in controlled experiments with low evidence ratio bags based on MNIST, as well as on a real life histopathology dataset - Camelyon16. Our method outperforms other methods in both bag level and instance level prediction, especially when only small training sets are available. We discuss the rationale behind our approach and the reasons for its superiority for these types of datasets.


\end{abstract}

\section{Introduction}

Multiple instance learning (MIL) is a form of weakly supervised learning where training instances 
are arranged in sets called bags, and a label is provided for the entire bag \cite{1} while the labels of the individual instances in the bag are not known. Weakly annotated data is especially common in medical imaging  \cite{2} where an image is typically described by a single label (e.g. benign/malignant) or a Region Of Interest (ROI) is roughly given.

We assume the case where a binary label is assigned to every bag in the dataset. The most common binary MIL assumption, is that the bag label is positive if at least one of the instances contains evidence for the label, and is negative if all of the instances do not contain evidence for the label. More formally, every bag is composed of a group of instances $\begin{Bmatrix} x_{1}, ...,x_{K} \end{Bmatrix}$, where $K$ is the size of the bag. $K$ can vary between the bags. A binary label $Y \in \begin{Bmatrix}0, 1\end{Bmatrix}$ is associated with every bag. The MIL assumption can then be written in this form: 

\begin{equation}
Y = \begin{Bmatrix}
 &0,& \textsl{iff } y_{k} = 0, \; for \; k \in \begin{Bmatrix} 1\dots 
 K \end{Bmatrix}\\ 
 &1,& otherwise 
\end{Bmatrix}
\label{eq1}
\end{equation}

In MIL a pooling operator is typically applied to aggregate the instance embeddings or predictions to create a bag output. Common choices for pooling operators are max-pooling Eq. \eqref{eq2}, or mean-pooling Eq. \eqref{eq3}.

\begin{equation}
Z_{m} = \max\limits_{k\in \{0 \dots K\}}(h_{km}),
\label{eq2}
\end{equation}
 
\begin{equation}
Z_{m} = \frac{1}{K}\sum_{k=1}^{K}h_{km},
\label{eq3}
\end{equation}
where $Z_{m}$ is the bag level prediction,  $h_{km}$ is the instance prediction value and $k\in \{0 \dots K\}$ and $m$ are the instance and bag indices respectively.

A more general formulation has been proposed in \cite{10} by assigning every instance a learned attention weight Eq. \eqref{eq4}.

\begin{equation}
Z_{m} = g(\sum_{k=1}^{K}a_{km}e_{km}),
\label{eq4}
\end{equation}
where $Z_{m}$ is the bag level prediction, $g$ is the bag level classifier, $e_{km}$ is the instance embedding,  $a_{km}$ is the instance attention value and $k\in \{0 \dots K\}$ and $m$ are the instance and bag indices respectively.

Recently there has been increased usage of MIL on large datasets, especially in the field of computational pathology \cite{10,18,19}. In the MIL setting for computational pathology, whole slide images (WSIs) are given a global label (e.g. "Tumor" if tumor cells exist in this WSI and "Normal" otherwise.). Instances are then extracted from the slides by sampling image tiles from the WSI with or without overlap. Instances (tiles) are then grouped into bags where every bag contains the tiles extracted from a specific slide and has that slide’s global label. 

In some cases just a small portion of the instances contains evidence for the global slide label, e.g. when the tumor is localized in a small part of the biopsy, which is usually the case. The interpretability of MIL algorithms is based on the ability to identify predictive instances in the bag \cite{10}. In addition, the sizes of the bags can be very large. For example, in the case of digital pathology, some slides in full resolution can have tens of thousands of instances extracted from them. These factors form a challenging MIL setting. As the bags grow larger, the ability of the network to correctly classify the bag is diminished and tightly linked to the selected pooling function. In the case of a low evidence ratio bag (i.e. a bag with a small number of positive instances compared to the total number of instances), if mean-pooling is used, a large negative instance population in the bag will overshadow the positive instances and create a false negative global bag level prediction. On the other hand, if max-pooling is used, a single negative instance with a high prediction value can corrupt the resulting global bag level prediction and create a false positive result. This is magnified by the unstable nature of deep learning models, where a small change in the input image can trigger a very different output \cite{8}. Given this setting, a large bag that contains many visually similar looking instances, might result in very different embeddings or predictions for each of them. Therefore pooling functions are a key element in any MIL algorithm.

In this paper we propose a novel pooling strategy for MIL that addresses the shortcomings of the current pooling functions and deals with the underperformance of MIL in the case of bags with low evidence ratio. We test and compare our method against baseline pooling methods i.e. max and mean pooling, as well as a state of the art MIL pooling method, namely Attention Pooling MIL. We conduct the algorithm tests and comparisons on both a controlled MNIST based dataset and a real life pathology image dataset - Camelyon16. Additionally, we explore the effect of dataset size on performance metrics of different MIL algorithms and evaluate performance on both datasets by examining bag level prediction and instance level prediction.

\section{Related Work}

In the context of digital pathology, an example of a recent use-case of MIL combined with deep learning for classification of prostate cancer Hematoxylin \& Eosin (H\&E) stained WSIs is described in \cite{3}. A huge dataset of 12,000 slides were extracted into 12,000 bags, with 1,000 instances per bag on average. Resnet18 pre-trained on Image Net was used to extract feature embeddings for the instances. Then a classification neural network with fully connected layers was trained on these embeddings using the max pooling operator (selecting the instance with the highest score for the cancer category).

The Attention Pooling MIL method \cite{10} is a state of the art MIL algorithm that uses a neural network to assign an attention weight score for every instance. These weights can then be used to aggregate the embeddings of the instances into a global slide embedding (by multiplying every embedding by its weight, and then summing over all embeddings). In the MNIST based experiments presented in \cite{10}, the bags used have an evidence ratio of 10\%.

Measuring certainty in deep learning networks, by Monte-Carlo (MC) dropout was introduced in \cite{11}. In MC dropout, dropout is applied in test time, and a forward pass is performed multiple times in order to capture the certainty of the model predictions. In \cite{12} certainty was used for multi-task learning by weighting individual task losses to create a global loss function. During the training, tasks with lower certainty receive weaker gradients. 

 We propose an MIL approach that uses the MC-dropout Mean-STD method for certainty calculation and generates weaker gradients during training for instances the model is not sure about. To the best of our knowledge this is the first work describing using certainty in the context of MIL algorithms.

\section{Proposed Method}

We formulate a new certainty based pooling function, which we call Certainty Pooling, that aggregates over the bag instances using the certainty score of the individual instances. We define $X_{k}$ to be the vector of MC dropout predictions for instance $k$. In Eq. \ref{eq5} we define the instance certainty $C_{k}$ as the inverse standard deviation of $X_{k}$.

\begin{equation}
C_{k} = C(X_{k}) = \frac{1}{{\sigma(X_{k})} + \epsilon},
\label{eq5}
\end{equation}
where $\sigma$ is the standard deviation operator and $\epsilon$ is a small number that prevents division by zero.

In Certainty Pooling Eq. \eqref{eq6}, we define the global bag level prediction $Z_{m}$ as the prediction value of the instance having the highest certainty weighted model output.

\begin{equation}
Z_{m} = h_{k^{*}m} \;\; where \;\; k^{*} = argmax(C_{k}\cdot h_{km}),
\label{eq6}
\end{equation}
where $k^{*}$ is the index of the instance having the highest certainty weighted model output.

\begin{figure}[t]
\centering
\includegraphics[width=1.0\textwidth]{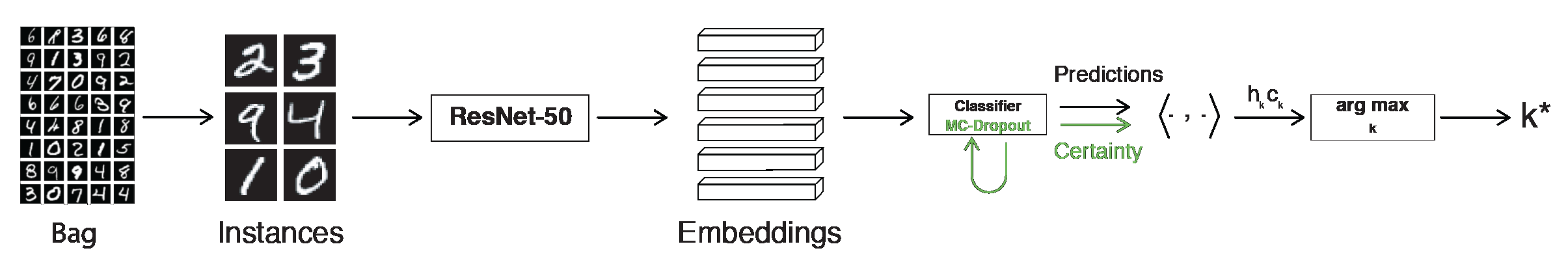}
\caption{Visual representation of the proposed method and model architecture.}
\label{fig_model}
\end{figure}

Additionally, a visual representation of the proposed method and model architecture is presented in Fig. \ref{fig_model}. 

\section{Experiments}
\subsection{Low Evidence Ratio MNIST-Bags}
The aim of this experiment is to test our method in a controlled dataset scenario where the bag evidence ratio is known and to compare it with current baseline and state of the art MIL methods. In order to make a fair comparison, we use the MNIST-Bags dataset proposed in \cite{10}. In this dataset, the bags are made of instances that are MNIST digits, where the bags can have a varying size. Bags containing the digit 9 are labeled as positive bags and bags without the digit 9 are labeled negative. In the originally proposed dataset bag evidence ratio was 10\%, i.e. on average 10\% of the instances in the positive bags are 9. With a  dataset size of 200 bags and the setting used in \cite{10}, both the Attention Pooling MIL and our method achieve near 100\% AUC.

In order to test and compare the behavior of the methods in low bag evidence ratio scenarios we explore a similar but more challenging dataset in which the evidence ratio is only 1\%. We select a constant bag size of 100 instances and define positivity in the same way. Positive bags contain exactly a single instance 9, and negative bags do not contain any instances of 9.  We explore the effect of the number of bags in the training set on the performance for the different methods. We use a validation set of 1000 bags and test our models on a testing dataset of 1000 bags. In all training and validation datasets, as well as in the testing dataset, exactly one half of the bags are positive.

We use the same network architectures as defined in \cite{10} and only replace the attention network with the proposed certainty based calculation.

In order to account for the stochastic nature of deep neural network (DNN) convergence, we repeat the experiment 20 times with different random seeds for every algorithm and  training set size, and take the average of the top K=10 results. The best model is selected for each method based on its performance on the validation dataset.

As a metric, we use the AUC value. We measure and compare the AUC of the bag level prediction, as well as instance level prediction on the testing set for the different sizes of training sets in each of the tested methods.

\begin{figure}[t]
\centering
\includegraphics[width=0.7\textwidth]{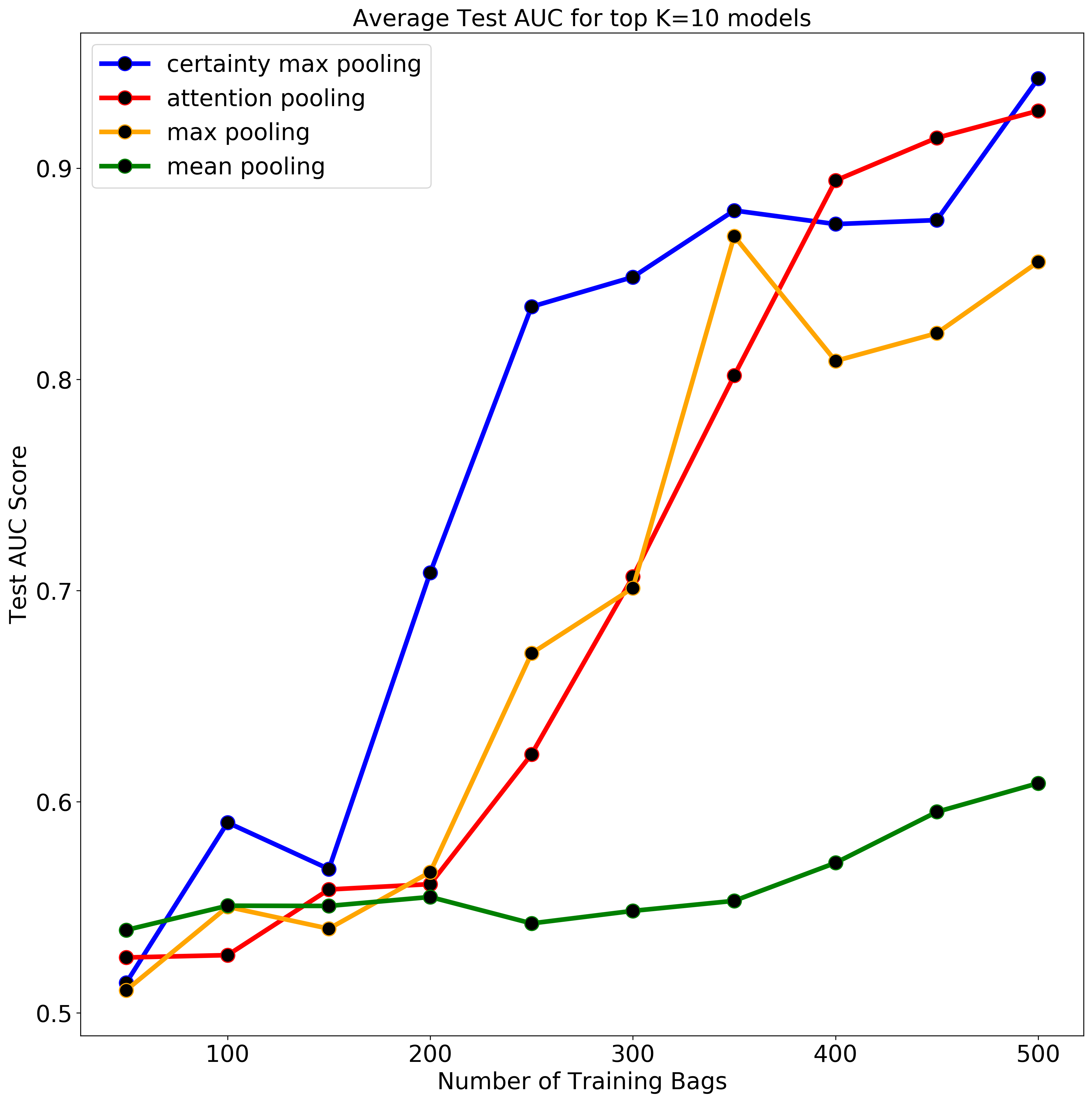}
\caption{Testing dataset bag prediction AUCs for models trained with different sizes of low evidence ratio MNIST bag datasets. In all experiments we consider the average of the top K=10 runs.
}
\label{fig1}
\end{figure}

The performance of our method compared to the benchmark methods on a low evidence ratio task with different sizes of training datasets is shown in Fig. \ref{fig1}. The graph shows how our method outperforms classical pooling methods as well as the benchmark Attention Pooling MIL method in most cases, and especially in small dataset sizes.

Some insights regarding the nature of stochasticity of these methods can be seen in Fig. \ref{fig1}. We find that the comparison between the Attention Pooling MIL and the Certainty Pooling MIL results is especially revealing. One can observe that with a very low number of bags in training set (50-150) both methods are rarely able to generate a good model for the testing set. In the interval between 200 and 300 training set bags Certainty Pooling seems to have more and higher results compared to the other methods. With training set sizes of 350 to 500 bags, Attention Pooling MIL results become comparable and even slightly better than Certainty Pooling. These observations could mean that in the case of low evidence ratio datasets where not enough data is available, Certainty Pooling can provide a better model where other methods encounter difficulties.

\begin{figure}[t]
\centering
\includegraphics[width=0.7\textwidth]{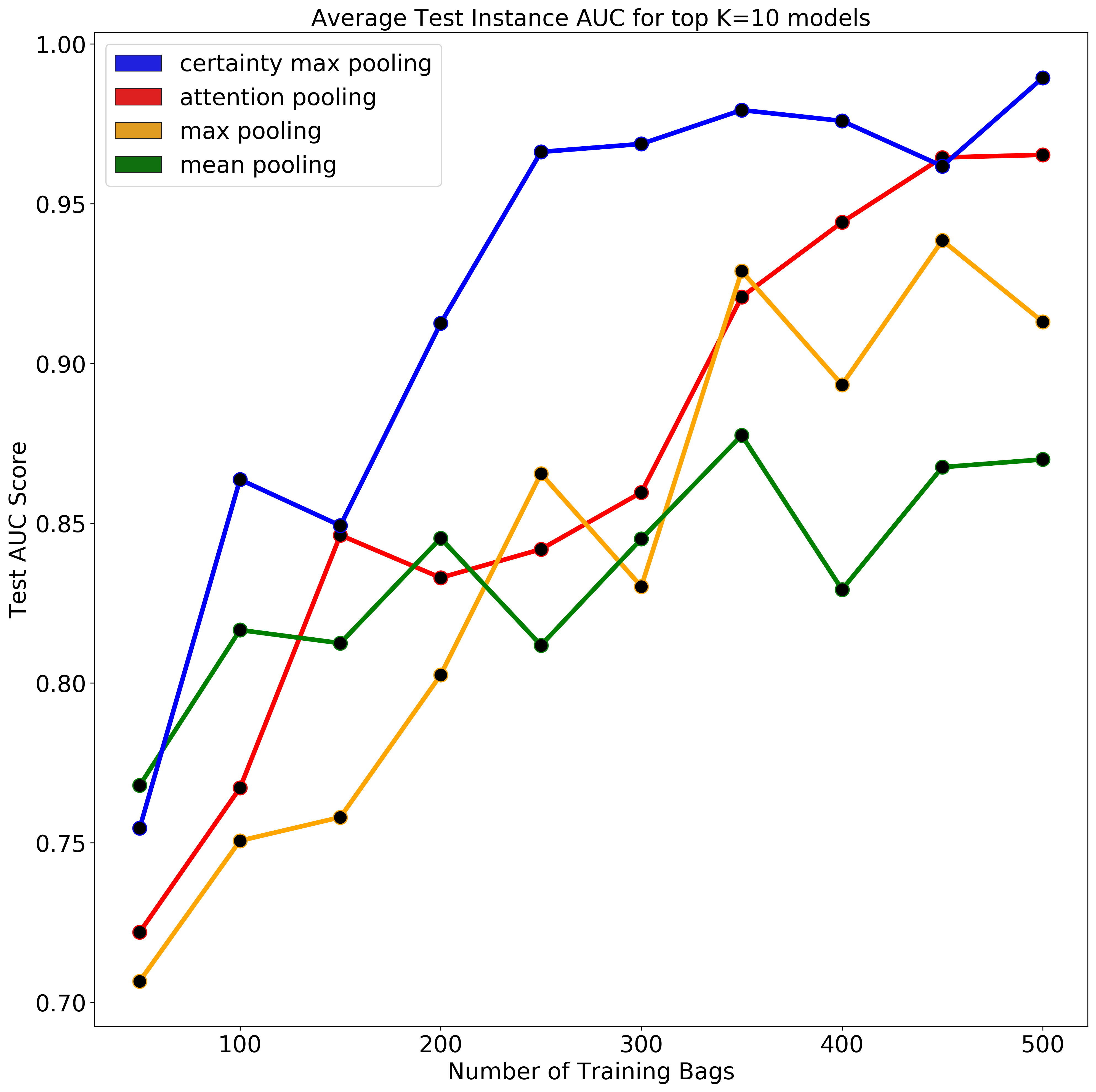}
\caption{ Average instance prediction AUC values for the top K=10 models.}
\label{fig3}
\end{figure}

Instance level prediction AUC is a practical way to assess the performance of the trained classifier on the instance level. This metric is calculated by using the instance level labels and predictions. This enables a comparison at test time of the different methods without dependence on the MIL pooling methods and shows how well the MIL training was able to learn a meaningful representation of the positive instances. In Fig. \ref{fig3}, we show the average instance prediction AUC values for the top K=10 models for all the test dataset instances based on the instance level labels (e.g. label is positive if instance is the digit 9 and negative otherwise). It can be observed that our method yields better results in both instance level prediction as well as bag level prediction. Given that all the compared methods were using the exact same network architecture, this might imply that training with Certainty Pooling provides more meaningful gradients during training and is therefore able to train a better instance level model predictor.

\begin{figure}[t]
\centering
\includegraphics[width=0.48\textwidth]{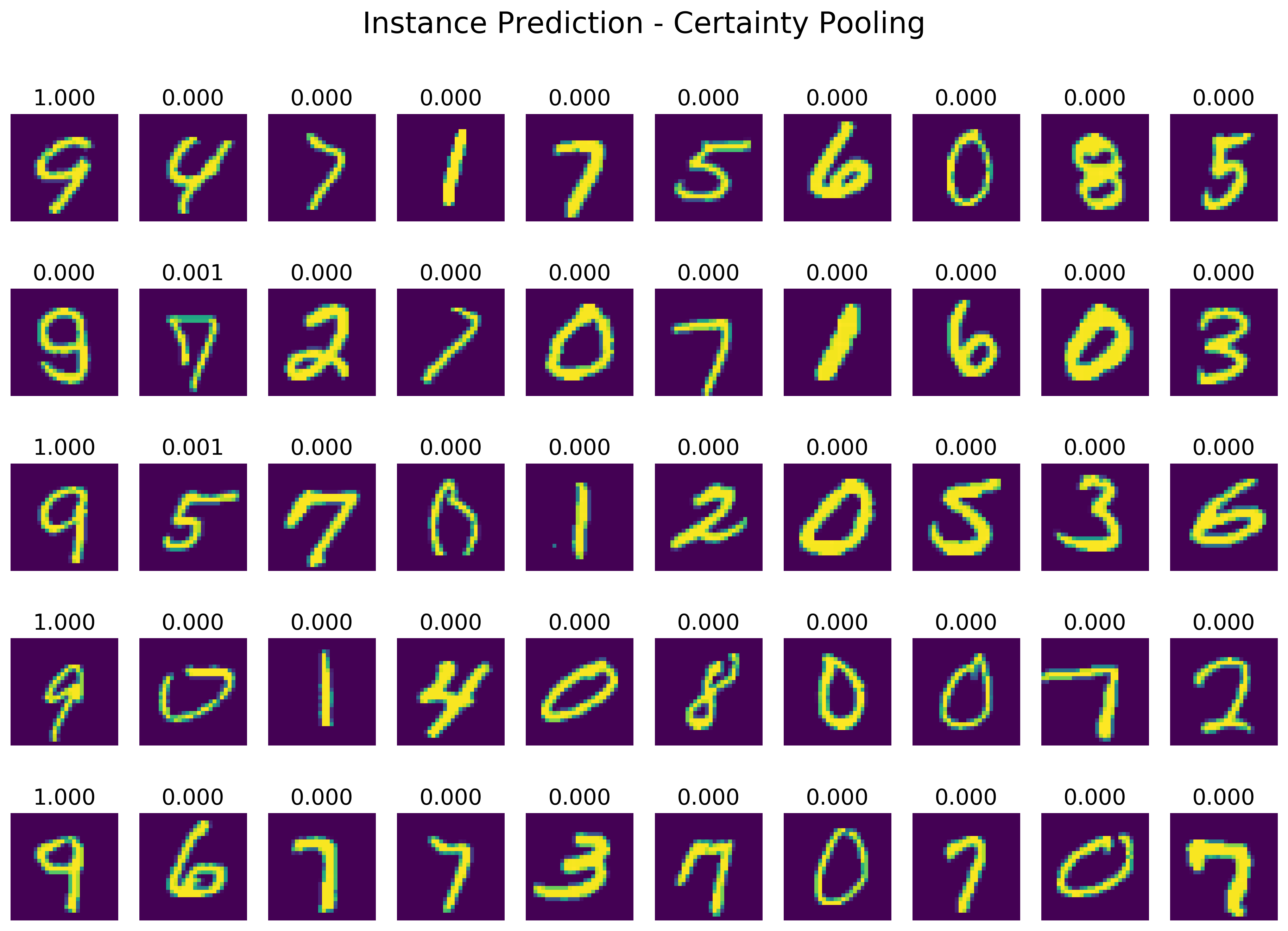}
\hfill
\includegraphics[width=0.48\textwidth]{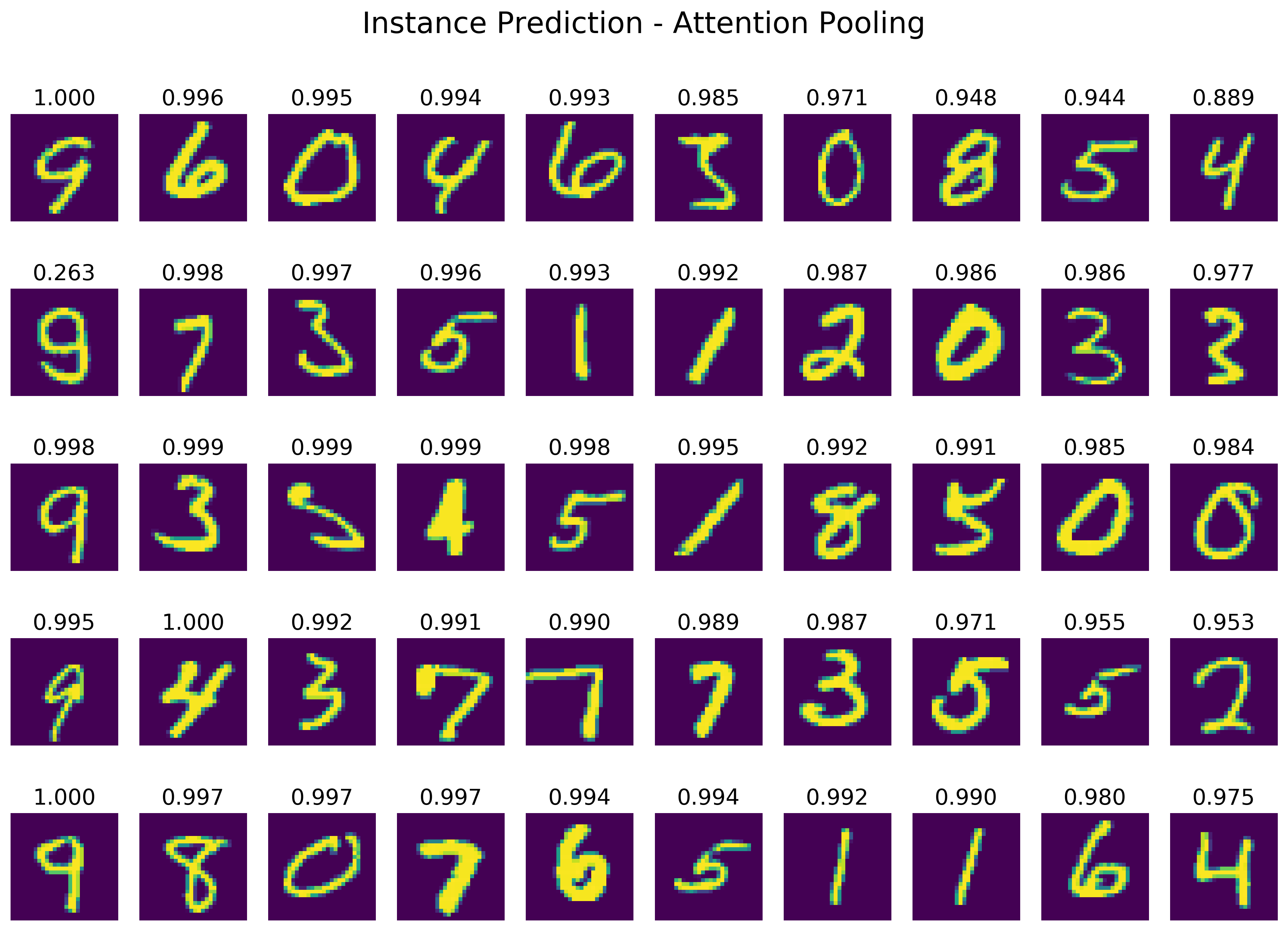}
\caption{Example instance level prediction values for Certainty Pooling MIL (left) and Attention Pooling MIL (right). The positive instance followed by the highest value instances in the bag ordered in descending order are presented and the value above each image represents the prediction value for that instance.}
\label{fig4}
\end{figure}

In Fig. \ref{fig4}, we show an example visualization of the instance prediction values assigned to different instances in selected bags for the Attention Pooling MIL method and the Certainty Pooling method based on one of the conducted experiments with a middle range training dataset size of 300 bags. For each method we choose the model parameters based on the best validation score from all experiments run on the selected dataset size. The first digit in each row is a 9 which is the positive instance, while the other instances are negative instances presented in descending order of their prediction value. We can observe that the differences between the prediction value for the positive and negative instances are  much larger with the Certainty Pooling MIL than the  Attention Pooling MIL algorithm, demonstrating the ability of our method to better train a classifier to select key instances.

\begin{figure}[t]
\centering
\includegraphics[width=0.49\textwidth]{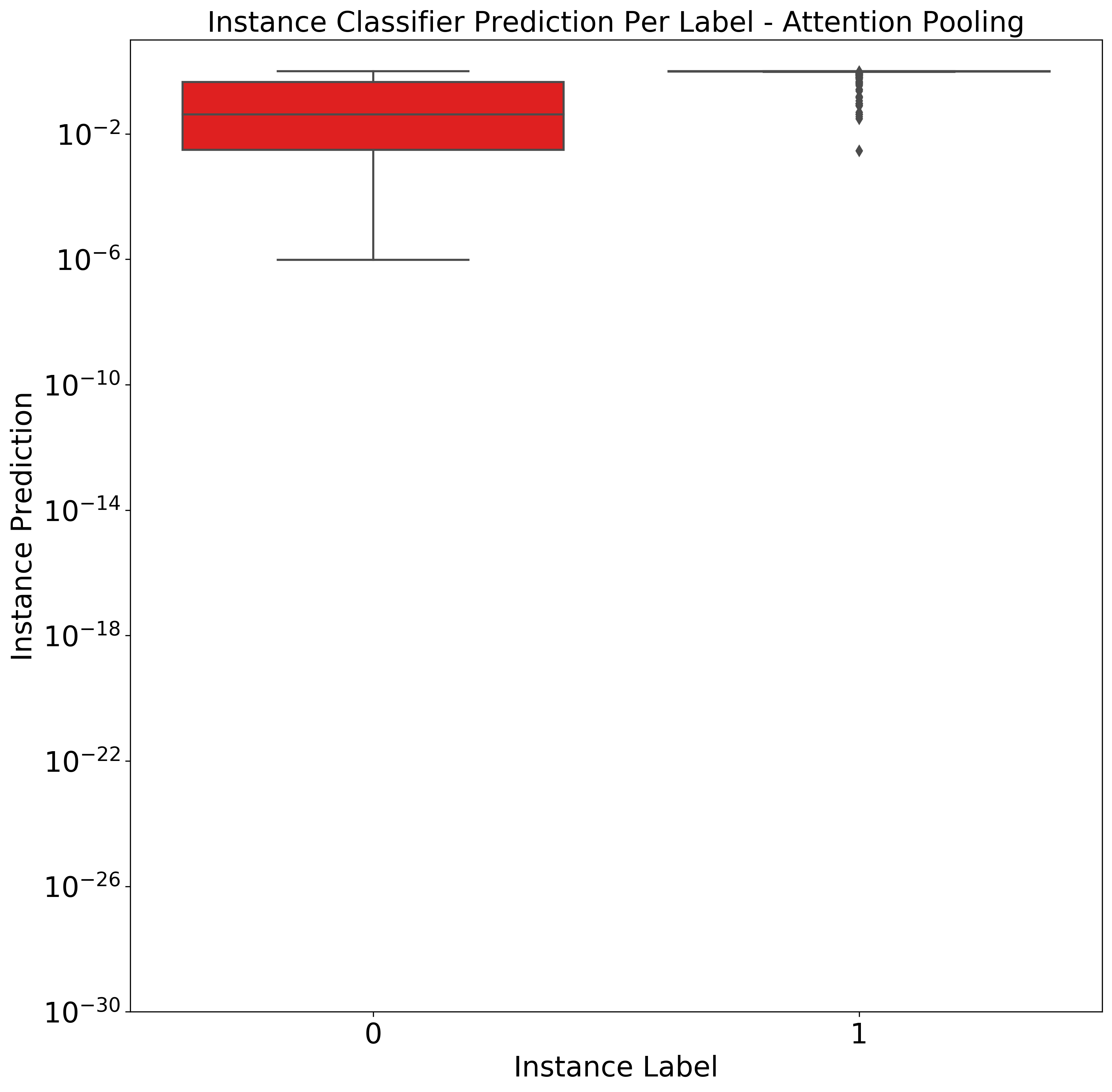}
\hfill
\includegraphics[width=0.49\textwidth]{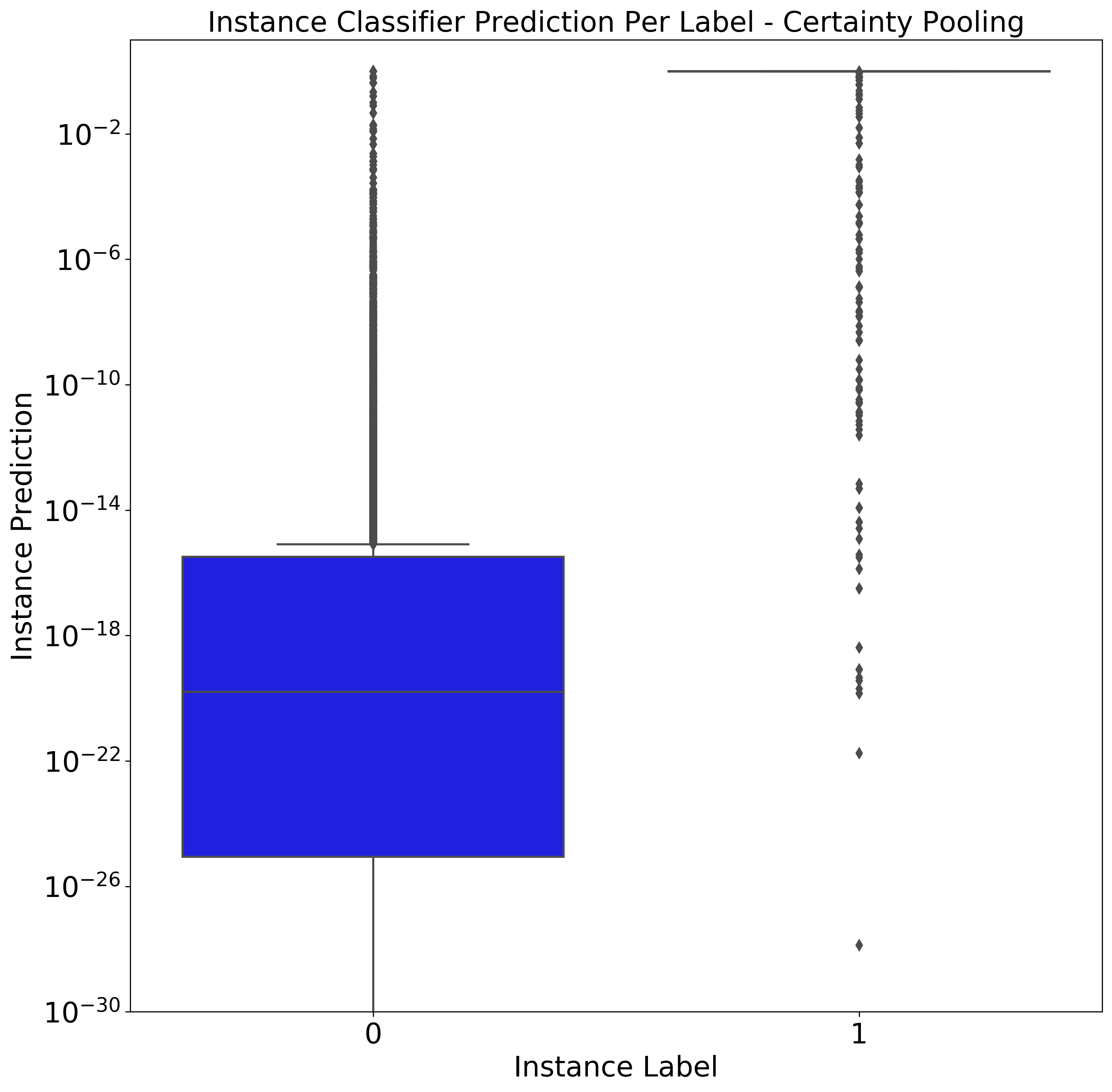}
\caption{Distribution of instance level prediction values for Attention Pooling MIL (left) and Certainty Pooling MIL (right) on the entire MNIST-Bags testing dataset.}
\label{fig41}
\end{figure}

Fig. \ref{fig41} shows the distribution of instance predictive values for the entire MNIST-Bags testing dataset for Certainty Pooling vs. Attention Pooling. The graph further illustrates how for Certainty Pooling, the instance prediction values for the negative and positive labels are much better separated than for Attention Pooling. This clarifies the better instance prediction AUC values observed for Certainty Pooling in comparison to the other methods, as can be seen in Fig. \ref{fig3}.

\subsection{Camelyon16 Lymph Node Metastasis Detection Challenge}

In this experiment we evaluate and compare our proposed method to other MIL methods on a challenging real life dataset. The Camelyon16 dataset consists of 400 H\&E stained WSIs taken from sentinel lymph nodes, which are either healthy or exhibit metastases of some form. In addition to the WSIs, the dataset contains both slide level annotations, i.e. healthy or contains-metastases and pixel level segmentation masks per slide denoting the metastases. 

We use the 270 WSIs in the Camelyon16 training set for training and 130 WSIs from the Camelyon16 test set for testing our algorithm. We investigate the effect of dataset sizes on algorithm performance in the real-life dataset paradigm, we conduct multiple experiments, each with a different training set size. In order to do this, we randomly select a fraction of the training dataset slides in 10\% increments between 50\% and 100\% and use only that selection for model training. From the selected dataset slides we holdout 25\% of the slides as a validation set, and use only the global slide labels for training. 

We extracted 256x256 non-overlapping tiles at 20x resolution which is the working magnification used by clinical pathologists to review slides. Simple image processing based tissue detection was applied to discard background (white) tiles. The tiles in each slide were grouped into a bag, where every bag contains only tiles from one specific WSI. In Histology applications, a step called stain normalization is typically applied to normalize tissue staining in different slides \cite{20,21}. We apply a simple stain normalization step, by normalizing the mean and standard deviation of the LAB color space channels to be the same as in a reference image from the Camelyon16 training set \cite{22}. 

The resulting training/validation dataset contains roughly 4 million instances (tiles) in 270 bags (slides). Similar to \cite{3} we avoid learning directly on the instance images because of the computational cost and instead we first extract 2048 length features for every instance using Resnet50 \cite{13} pre-trained on ImageNet.

The prediction network we used consists of 5 fully connected (FC) layers. The first hidden layer has 1024 neurons, and every layer following has half the amount of the previous. We introduce a dropout layer with  50\% dropout after each FC layer and ReLU activation. We set the learning rate to 0.01. For Attention Pooling MIL we used the attention network architecture suggested in \cite{10}, but increased the number of neurons to 1024 in the hidden layers, to fit the higher complexity of this data. In all cases, an Adam optimizer was used with default parameters.

Similar to \cite{16}, we randomly select 128 instances from every bag during training in each epoch as an augmentation strategy. We found this largely improved results and prevented quick over-fitting on the training set. We train the different models for 1000 epochs and selected a model that performed best on a held out validation set. We test the selected model on the Camelyon16 test set tiled and generated in a similar fashion. During testing, no instance sampling is performed and the full bag is analysed. As described in the competition instructions, we report the AUC metric on the Camelyon16 test set for the global labels. 

\begin{figure}[t]
\centering
\includegraphics[width=0.7\textwidth]{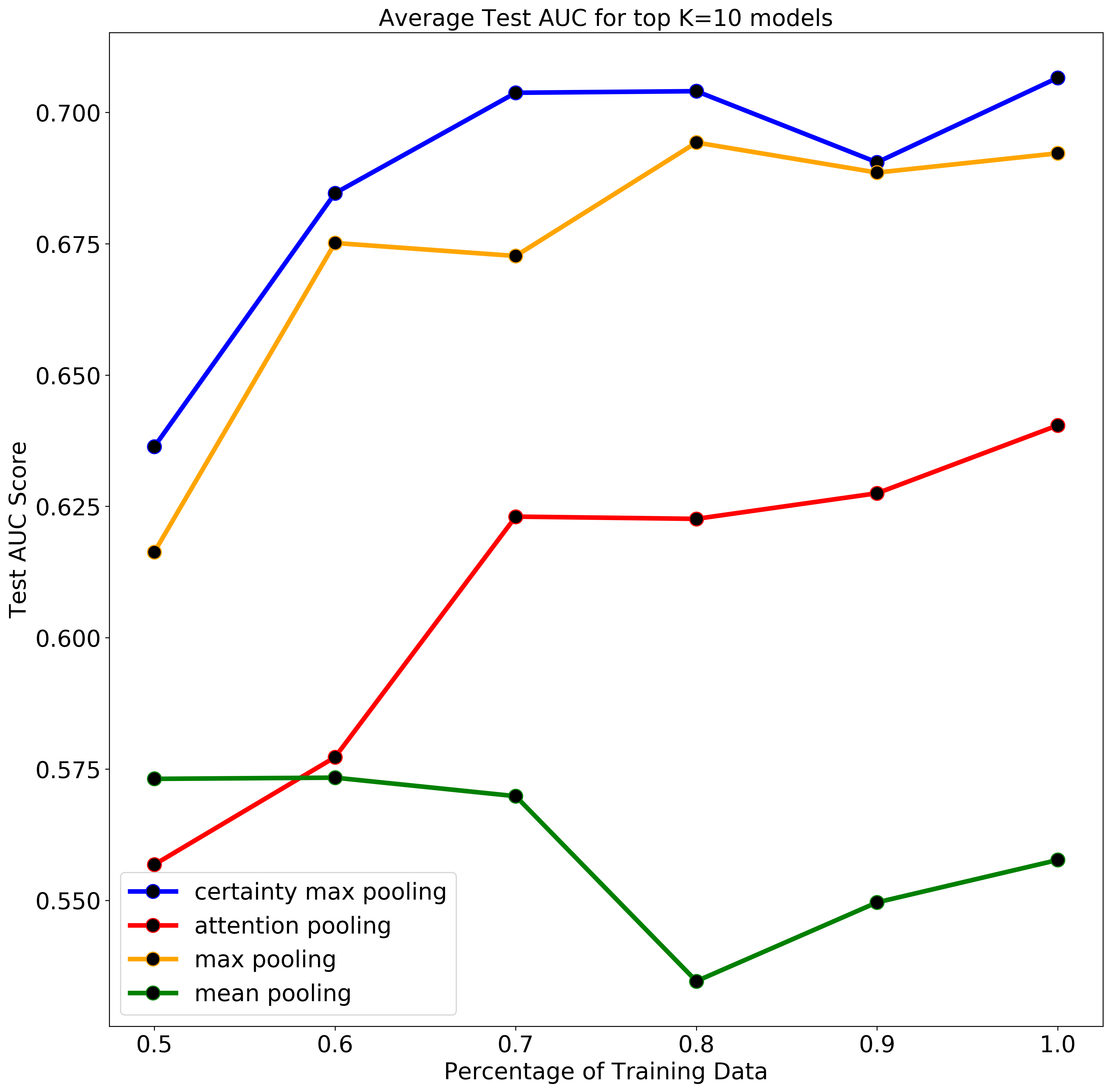}
\caption{Camelyeon16 testing set average top K=10 global bag prediction AUCs for different training dataset sizes.}
\label{fig_camelyon_bag}
\end{figure}

In order to account for the stochastic nature of deep neural network (DNN) convergence, we repeat each experiment 20 times with different random seeds for every algorithm and training set size. We try to reduce the prohibitively long duration of the multiple seed experiments by checking the validation set AUC score, every 5 epochs and randomly sampling 20,000 tiles from each slide in the validation set to compute the slide prediction. We run the training for 1000 epochs and select the best model for each method based on its performance on the validation dataset. The average of the top K=10 results is presented in Fig. \ref{fig_camelyon_bag}.  Our method achieves top bag level AUC results in all datapoints compared to both the baseline MIL pooling methods and the benchmark Attention Pooling MIL method.

Additionally we calculate the instance level prediction AUC for each method by using the Camelyon16 WSI mask annotations to label each tile. Due to the inherent huge imbalance of the instance labels, i.e. only 42 of the 137 slides are positive, and among these slides less than 6\% are positive instances. The instance level prediction AUC is computed for each positive slide separately and averaged across slides. instance level prediction AUC is defined only for positive slides since only they contain both positive and negative tiles. 

It is interesting to note that from the testing set, only 17 out of 42 positive slides have an evidence ratio higher than 1\% and only 7 above 5\%, meaning the majority of test set slides have an evidence ratio of much less than 1\%. 

\begin{figure}[t]
\centering
\includegraphics[width=0.7\textwidth]{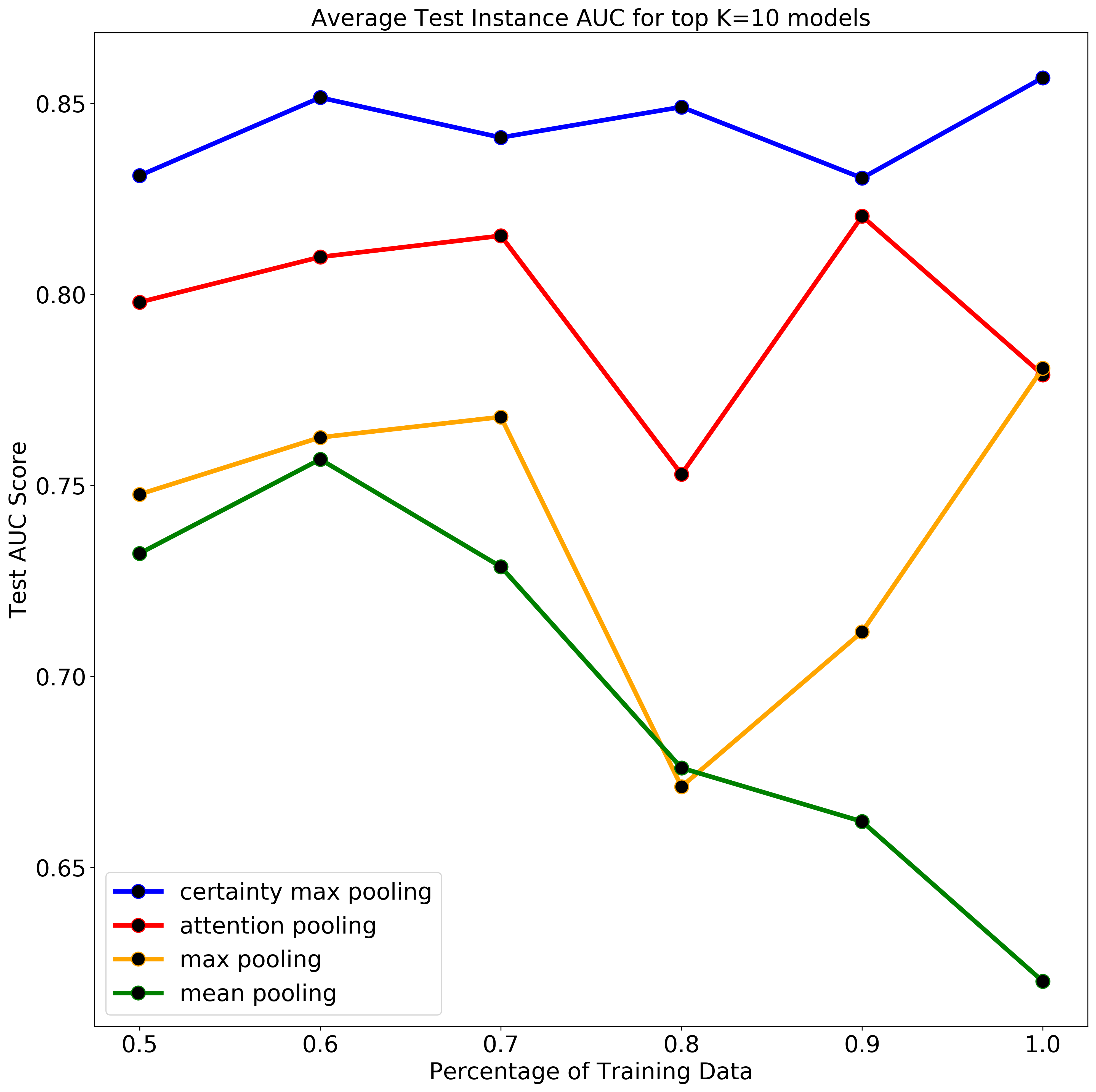}
\caption{Camelyeon16 testing set average top K=10 global instance prediction AUCs for different training dataset sizes.}
\label{fig_camelyon_instance}
\end{figure}

The average top K=10 instance level prediction AUC per method on the testing set are displayed in Fig. \ref{fig_camelyon_instance}. Our method achieves top instance AUC results in all datapoints compared to both the baseline MIL pooling methods and the benchmark Attention Pooling MIL method.

\begin{figure}[t]
\centering
\includegraphics[width=0.49\textwidth]{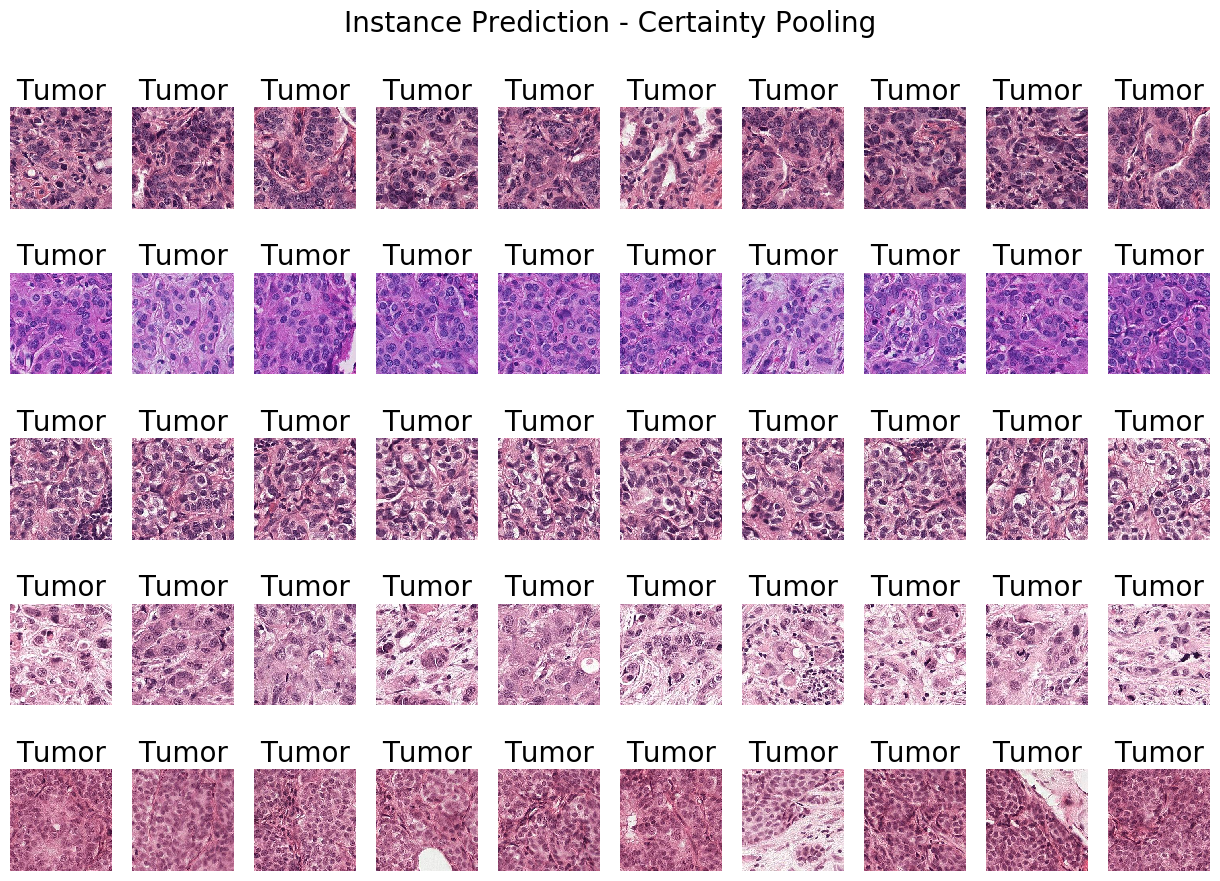}
\hfill
\includegraphics[width=0.49\textwidth]{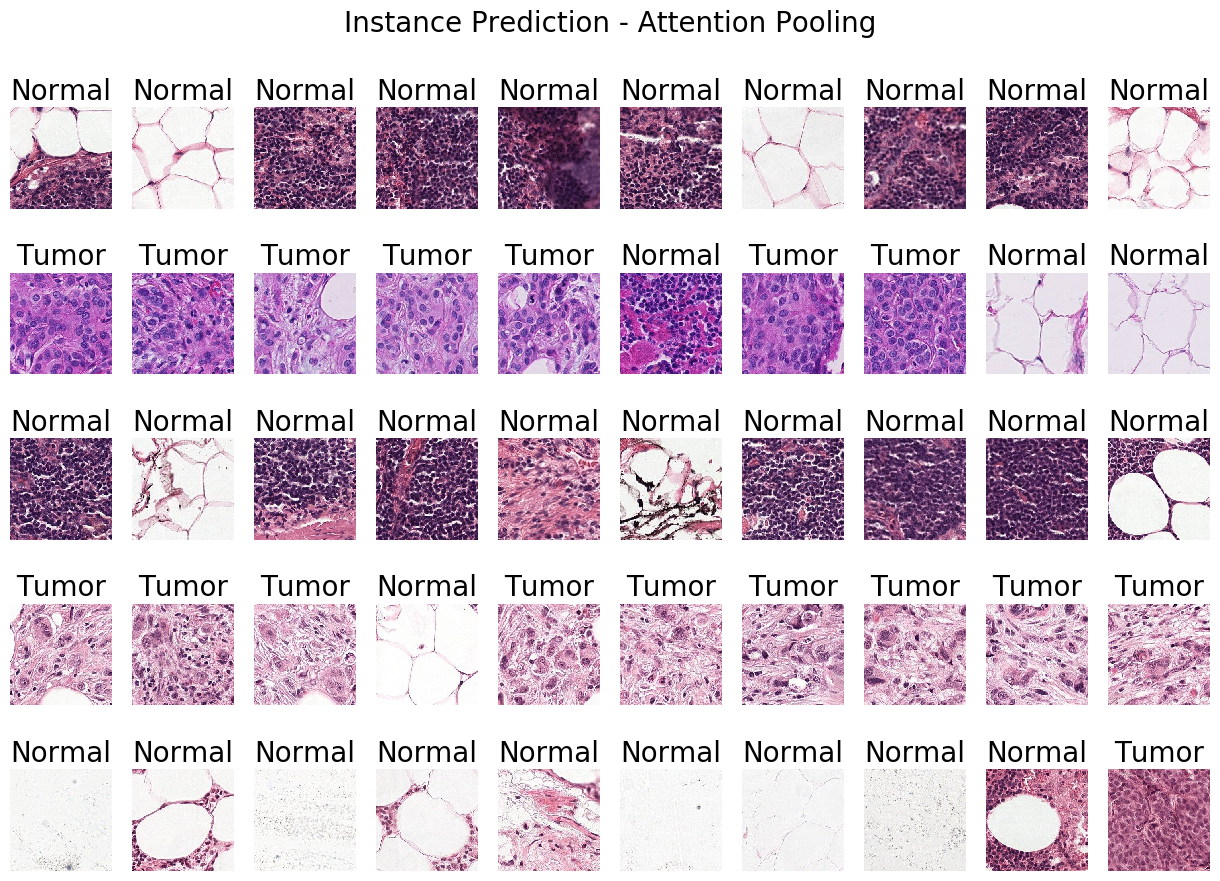}
\caption{Example instance level prediction tiles for Certainty Pooling MIL (left) and Attention Pooling MIL (right). The top instances sorted by predicted instance (tile) value for 5 "Tumor" labeled slides. Above each tile the instance level label from the Camelyon16 mask labels.}
\label{fig5}
\end{figure}

In Fig. \ref{fig5} a gallery of top predicted instances for 5 positive (Tumor labeled) slides from the Camelyon16 testing set are presented for Certainty Pooling MIL and Attention-MIL. It is visible in the galleries that while Certainty Pooling produces a classifier that retrieves only tumor labeled tiles in the top-10 tiles per slide for the example slides, the Attention-MIL trained classifier does not perform as well on the instance level.

\section{Conclusions}

In this paper, we introduce a novel MIL pooling function based on network certainty measures. Our method is based on calculation of the instance classifier network certainty and then weighting instance predictions by their associated certainty.

We test our methods on a "controlled" MNIST based dataset which is a modification of the MNIST-Bags dataset that creates a very challenging setting, which we call "Low Evidence Ratio MNIST-Bags", as well as on the  Camelyon16 "real-life" histology dataset which has over 4 million instances and a low evidence ratio. Certainty Pooling improves both Bag Level Prediction and Instance Level Prediction in both experiments.

We argue our method is able to train a more interpretable model in terms of key instance retrieval. We demonstrate this by looking at top prediction instances and show that our method presents more relevant instances with high value prediction. Our method also improves results in MIL settings where the ratio of instances containing evidence is low. As MIL is increasingly being applied on challenging datasets with limited size, we believe the improvements presented can have a real impact on the quality and interpretability of trained MIL networks.

We rationalize the approach of choosing certainty as a weighing factor for the pooling operation as follows. First, we argue that the learned model tends to be more certain about predictive instances than about non-predictive instances. One possible explanation for this is that since non-predictive instances appear both in positive and negative bags (e.g. digits 0-8 in MNIST and normal tissue or background tiles in Camelyon16), they receive contradicting gradients during training, causing a divergence in the network trained with Dropout layers. This in turn creates fluctuations in the predicted instance value during inference with Dropout and causes a large variance in the instance prediction values. We interpret this variance as model uncertainty in the prediction of this instance and use it to weigh the contribution of the instance on the bag prediction. 

We think it is natural to decrease the contribution of instances the model is not sure about, especially in the scenario of large bags where there are many opportunities for mistakes that can affect the bag prediction. When the instance attention levels are directly learned as with Attention Pooling MIL, there is no direct control over which instances will achieve high attention values. 

Our method provides a simple way to dynamically weigh instances in an explainable way without learning the weight via another neural network. By doing this we believe our method bridges the gap between traditional mean and max pooling operators and more advanced learned weighting mechanisms such as attention.

Future work will include hyper-parameter optimizations, working image resolution and resolution combinations as well as investigation into the limits of dataset sizes and evidence ratio for training certainty based MIL algorithms. We also plan to investigate the advantages of more sophisticated color normalization schemes in this scenario.

\bibliographystyle{splncs04}
\bibliography{eccv2020submission}
\end{document}